\documentclass{ivcnz}
\usepackage{graphicx}
\usepackage{amsmath,amssymb} 
\usepackage{color}
\usepackage{balance}
\usepackage[pagebackref=false,breaklinks=true,colorlinks,bookmarks=false,urlcolor=blue,linkcolor=blue,citecolor=blue,pdftitle={Video Face Matching using Subset Selection and Clustering of Probabilistic Multi-Region Histograms},pdfauthor={Sandra Mau, Shaokang Chen, Conrad Sanderson, Brian C. Lovell}]{hyperref}

\usepackage[british]{babel}

\def\Vec#1{{\mathbf{#1}}}

\title
  {
  Video Face Matching using
  Subset Selection and Clustering
  of Probabilistic Multi-Region Histograms
  }

\author{Sandra Mau, Shaokang Chen, Conrad Sanderson, Brian C. Lovell}

\address
  {
  NICTA, PO Box 6020, St Lucia, QLD 4067, Australia\\
  The University of Queensland, Brisbane, QLD 4072, Australia
  }

\begin{document}

\begin{abstract}

Balancing computational efficiency with recognition accuracy is one of the major challenges in real-world video-based face recognition.
A significant design decision for any such system is whether to process and use all possible faces detected over the video frames,
or whether to select only a few `best' faces.
This paper presents a video face recognition system based on probabilistic Multi-Region Histograms
to characterise performance trade-offs in:
(i)~selecting a subset of faces compared to using all faces,
and
(ii)~combining information from all faces via clustering.
Three face selection metrics are evaluated for choosing a subset:
face detection confidence, random subset, and sequential selection.
Experiments on the recently introduced \mbox{MOBIO} dataset
indicate that the usage of all faces through clustering
always outperformed selecting only a subset of faces.
The experiments also show that the face selection metric based on face detection confidence
generally provides better recognition performance than random or sequential sampling.
Moreover, the optimal number of faces varies drastically across selection metric and subsets of \mbox{MOBIO}.
Given the trade-offs between computational effort, recognition accuracy and robustness,
it is recommended that face feature clustering would be most advantageous in batch processing
(particularly for video-based watchlists),
whereas face selection methods should be limited to applications with significant computational restrictions.

\end{abstract}

\keywords{surveillance, local features, video processing, face matching, clustering, subset selection, efficiency.\thanks{{\bf Published~in:} International Conference of Image and Vision Computing New Zealand (IVCNZ), 2010. \href{http://dx.doi.org/10.1109/IVCNZ.2010.6148860}{http://dx.doi.org/10.1109/IVCNZ.2010.6148860}}}%
\maketitle

\section{Introduction}
\label{sec:intro}

While there has been a substantial amount of research in still image face recognition,
there has been comparatively less on video face recognition.
Video typically provides much more information for recognition compared to still images,
including temporal and multiview information.
However, one of the major challenges in video is to decide how to maximise the usage of available information
while ensuring the system can still run in a scalable and timely manner.
For example, despite typically having many frames of face information available from video,
one of the design decisions for any face recognition system includes how many faces to use and, if not all, how to select them.
There is potentially a trade-off between computational effort and recognition accuracy, which can be influenced by the number of faces used.

Additionally, we are interested in addressing real-world video recognition problems
where the environment is uncontrolled and subjects may not be actively cooperating with the camera.
Furthermore, the quality of images can vary quite dramatically.
For instance, in surveillance contexts, CCTV video suffers from
low quality, resolution mismatches, varying pose and lighting from camera to camera,
and also within the same camera depending on time of day (changes in lighting and shadows).
Another example of an uncontrolled environment is handheld mobile video,
which often suffers from quality issues such as lens smudging, blur, pose and lighting changes
due to variation between scenes (indoor/outdoor).
In addition to the above image variations, face detection and alignment
will also have great influence on the recognition performance~\cite{Gary-ICCV-2006}.
Many face recognition algorithms assume the faces are well aligned and normalised,
which may not be the case, especially for low quality video.
Thus to address these issues,
not only does the face recognition system need to be scalable and efficient,
but it also has to be robust to common issues that affect recognition accuracy.

This paper describes a system for video-to-video face recognition
which uses an adapted form of the probabilistic Multi-Region Histogram (MRH) method
originally developed for still-to-still face recognition~\cite{Sanderson2009ICB}.
We have chosen to extend it to video-to-video recognition as it has shown robustness to alignment errors
as well as variations in illumination, pose and image quality.
Furthermore, MRH is relatively computationally efficient,
making it suitable as a starting point for developing a scalable video-to-video recognition system.

Within the video-based system, we characterise the impact on recognition accuracy
when using various methods of face selection to choose only a subset of faces for recognition.
We also contrast those methods with the alternative of using information from all faces through feature clustering.
We examine the trade-offs between computational effort and recognition performance present in
face selection and clustering,
and suggest situations where the two approaches might be best utilised.

The paper proceeds as follows:
Sections~\ref{sec:prev_work} and~\ref{sec:prev_work_clustering} provide background on face selection and feature clustering in video;
Section~\ref{sec:algorithms} describes our video-based face recognition framework;
Section~\ref{sec:experiments} discusses the experiments on the \mbox{MOBIO} dataset to compare the different approaches in face selection,
and contrasts that to the utilisation of all faces through feature clustering.
Conclusions and directions for future work are given in Section~\ref{sec:conclusion}.

\vspace{-0.5ex}
\section{Background: Face Selection}
\label{sec:prev_work}
\vspace{-1ex}

While there has been several surveys on video-based face recognition~\cite{Matta2009JVLC,Wang2009WASET,Zhao03ACMCS,Shan10SCI},
face selection has not been reviewed as a component in existing face recognition systems until recently (2010)~\cite{Shan10SCI}.


As larger video datasets are being made available
including the Mobile Biometrics (\mbox{MOBIO})
\linebreak
dataset~\cite{MOBIO2010ICPR},
which has made face selection a more prominent topic to investigate.
MOBIO has 17,480 videos and over 3 million frames
--- with such a large amount of information,
balancing computational efficiency with recognition performance becomes very necessary.

%

Shan~\cite{Shan10SCI} calls the approach of independently using all or a subset of face images with a still image-based recognition method the `key-frame (or exemplar) based approach'. 
In most cases ad-hoc heuristics are used to select key-frames.
%
%
A common way of selecting a subset of faces is through a metric based on face detection confidence
after the face detection step~\cite{MOBIO2010ICPR}.
Face confidence metrics can be based on located facial features (such as eyes and nose) within the face~\cite{Gorodnichy02FG},
or face classification using pre-trained binary classifiers~\cite{Villegas08CVPR,Berrani05AVSS}.
The number of selected faces is typically chosen in a heuristic manner,
such as the number of faces or faces above a certain threshold of confidence.

There are typically two main reasons for not using all faces: 
the first is computational effort due to the size of the dataset,
the second is that the marginal gain in recognition accuracy decreases after a certain number of faces~\cite{MOBIO2010ICPR}.
We will discuss the computational effort trade-offs in Section~\ref{subsec:selection_clustering_compared}, 
and our experiments in Section~\ref{subsec:face_selection} will analyse the second reason in more detail.

\vspace{-0.5ex}
\section{Background: Face Clustering}
\label{sec:prev_work_clustering}
\vspace{-1ex}

In the cases where computation time is less of a limitation,
such as offline or batch processing,
there is potential to utilise information from all faces in a video.
However, using all faces for recognition must still be done in a tractable manner,
either in the recognition step itself or in a pre-processing step.
We propose to use facial feature clustering as such a pre-processing step.

Historically, video face recognition methods originate from still-image based techniques, 
which get applied over the multiple face frames by treating each as a still image~\cite{Zhao03ACMCS}
and modify the distance calculation to accommodate multiple identification hypotheses.
These approaches are classified by
\linebreak
Matta and Dugelay~\cite{Matta2009JVLC} as approaches that neglect temporal information.
This class of video face recognition makes up the majority of the face recognition systems published~\cite{Zhao03ACMCS}.
They include video extensions of PCA, LDA, Active Appearance Models and Elastic Graph Matching.
The major drawback to these approaches is that they may become computationally intractable
to store and search for any significant amount of video.  
They also do not take advantage of the fact that sequential faces
may be very similar and thus may be grouped together to reduce redundancy.

Temporal model and image-set matching 
\linebreak
approaches address this issue by modeling the distribution of face images over time or by features~\cite{Shan10SCI}.                                  
These approaches tend to integrate the information expressed by all the face images into a single model. 

One such image-set matching solution is to cluster similar faces by feature similarity.
%
%
%
Lee et al.~\cite{Lee03CVPR} proposed to learn a low-dimensional manifold,
which is approximated by piecewise linear subspaces. 
To construct the representation, exemplars are first sampled from videos by finding frames with the largest
distance to each other corresponding to head pose changes in video, which are further clustered using K-means clustering. 
Each cluster models face appearance in nearby poses, represented by a linear subspace computed by PCA. 
Arandjelovic et al.~\cite{Arandjelovi05CVPR} model the face appearance distribution
as Gaussian Mixture Models (GMMs) on low-
\linebreak
dimensional manifolds.             
In further work~\cite{Arandjelovi06ICPR}, they derived a local manifold illumination invariant,
and formulated the face appearance distribution as a collection of Gaussian distributions corresponding to clusters obtained by \mbox{{\it k}-means}.

We propose a similar approach of clustering face features as a collection of Gaussians as a pre-cursor to face recognition for any system.
This will be demonstrated using Multi-Region Histogram features rather than the previously used local manifolds.

\section{Video-to-Video Matching}
\label{sec:algorithms}
\vspace{-0.5ex}

\begin{figure*}[!tb]
\centerline{\includegraphics[width=0.8\textwidth]{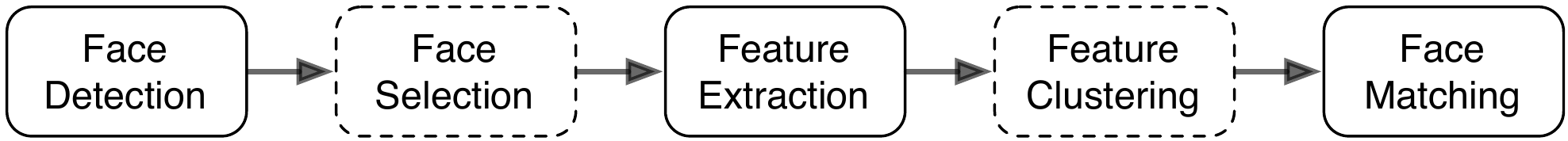}}
\vspace{-2ex}
\caption
  {
  \mbox{A recognition system with the proposed face selection and feature clustering steps highlighted.}
  }
\label{fig:flow}
\end{figure*}

A generic face recognition system has the components of face detection, feature extraction, and face matching~\cite{Zhao03ACMCS}.
The two system components being proposed and analysed in this paper, face selection and facial feature clustering,
fall in between the face detection and recognition steps, as illustrated in Fig.~\ref{fig:flow}.
The face recognition system presented here uses OpenCV for face detection
in conjunction with a modified form of MRH~\cite{Sanderson2009ICB} for feature extraction.
Details of the system components are given below.

%
%
%

\subsection{Face Localisation}
\vspace{-0.5ex}

For face localisation, OpenCV's Haar Feature-based Cascade Classifier~\cite{Viola2001IJCV}
is used to detect and localise faces in each frame.
Eyes are located within each face using a Haar-based classifier.
If no eyes are found, their locations are approximated based on the size of the localised face.
The faces are then resized and cropped such that the eyes are at predefined locations with a 32-pixel inter-eye distance.
The final face is a closely cropped `inner' faces of size 64$\times$64 pixels
(as later seen in Fig.~\ref{fig:kmeans_K002}),
which attempts to exclude image areas susceptible to disguises, such as the hair and chin.

\subsection{Face Selection}
\vspace{-0.5ex}


One approach for face selection is based on a metric of face detection confidence, 
which is the confidence of a face classifier that the region of interest is a face.
The implementation of this metric varies.
A generic method is to apply a post-processing step of a face or non-face binary classifier
for all faces detected to obtain a confidence measure~\cite{Villegas08CVPR}.
We compare a face confidence method in~\cite{MOBIO2010ICPR},
which is based on where landmarks are detected within the face (such as eyes and nose),
to more naive methods of random and sequential selection.

Given any video $V_i$ for person $i$, the number of faces extracted from the video is $N_i$.
Faces $l_{j}$ from video $V_i$ are sorted chronologically and indexed by $j$.
We can then select $m$ faces from $V_i$ to form a face set $S = \{l_{q_1}, l_{q_2}, \cdots, l_{q_m}\}$.
For random selection, $q_k = \operatorname{rand}(N_i), k \in [1,m]$,
where $\operatorname{rand}(N_i)$ generates a unique random number between $1$ and $N_i$.
For sequential selection, we select the first $m$ faces from $V_i$, that is $q_1 = 1, q_2 = 2, \cdots, q_m = m$.
For confidence selection, each face $l_j$ is processed by the face detector to get the confidence of the detection~$c_j$.
The top $m$ faces with the highest confidence are selected.

\subsection{Feature Extraction using MRH}
\vspace{-0.5ex}

The MRH approach is motivated by the concept of `visual words'
(originally used in image categorisation~\cite{Nowak_ECCV_2006})
as well as the semi-loose spatial constraints between face parts in 2D~Hidden Markov Models~\cite{Sanderson2009ICB}.
It can briefly described as follows.
A~given face is divided into several fixed and adjacent regions (e.g.~3$\times$3)
that are further divided into small overlapping blocks (with a size of 8$\times$8 pixels).
For region $r$ a set of low-dimensional feature vectors is obtained from the blocks in that region,
\mbox{\small $F_r = \{ \Vec{f}_{r,i} \}_{i=1}^{M}$}.
Each block is normalised to have zero mean and unit variance,
and  descriptive features are extracted from each block via 2D~DCT decomposition~\cite{Gonzales_2007}.
Each feature vector $\Vec{f}_{r,i}$ obtained from region $r$ is then represented as a high-dimensional probabilistic histogram:

\vspace{-2ex}
\begin{small}
\begin{equation}
   \Vec{h}_{r,i}
   \hspace{1pt}
   \mbox{=}
     \left[
        \frac{ w_1 ~ p_1\left(\Vec{f}_{r,i}\right)}{ \sum_{g=1}^{G} w_g p_g\left(\Vec{f}_{r,i}\right) },
        \hspace{-1pt}
       ~\cdots\hspace{-1pt},
       ~\frac{ w_G ~ p_G\left(\Vec{f}_{r,i}\right) }{ \sum_{g=1}^{G} w_g p_g\left(\Vec{f}_{r,i}\right) }
     \right]
\label{eqn:prob_histogram}
\end{equation}
\end{small}

\noindent
where the $g$-th element in {\small $\Vec{h}_{r,i}$} is the posterior probability of {\small $\Vec{f}_{r,i}$}
according to the $g$-th component of a `visual dictionary' model,
with an associated weight of $w_g$.
The dictionary is a Gaussian Mixture Model with 1024 components, built from low-dimensional 2D~DCT features extracted from training faces.
The mean of each Gaussian in the dictionary can be thought of as a particular `visual word'.
Robustness to face misalignment is achieved by representing each region as one average histogram:

\vspace{-2ex}
\begin{equation}
  \Vec{h}_{r,\mathtt{avg}} = \frac{1}{M} \sum\nolimits_{i=1}^{M} \Vec{h}_{r,i}
\label{eqn:region_avg}
\end{equation}

\noindent
For faces with a size of 64$\times$64 pixels, there are 9 regions arranged in a 3$\times$3 layout.
This results in an MRH signature composed of 9 histograms, with each histogram having 1024 components:

\vspace{-1ex}
\begin{equation}
  \operatorname{MRH} = \left[ ~h_{1,\mathtt{avg}}, ~h_{2,\mathtt{avg}}, ~\cdots, ~h_{9,\mathtt{avg}} ~\right]
  \label{MRH-whole}
\end{equation}

\noindent
One MRH signature is used to represent each face in each frame.

\subsection{Feature Clustering}
\label{subsec:feature_clustering}
\vspace{-0.5ex}

We choose the widely known $k$-means algorithm to group a set of faces into $k$ clusters
and represent each cluster by its centroid~\cite{Jain10PRL,Mackay2003}.
We adapt it to dealing with videos and MRH face signatures
by seeding the $k$ clusters with faces spaced at regular intervals within a video;
the distance metric used during the clustering process is described in Eqn.~(\ref{eqn:raw_dist}).

Once the $k$-MRH clusters have been generated, the average MRH of each cluster's signatures is used as the representative signature.
The special case of $k=1$ is just an average MRH signature over all available faces.
In the experiments we also apply clustering on faces from multiple videos belonging to the same person.


%


%
%
%

\subsection{MRH Signature Comparison}
\label{subsec:face_rec}
\vspace{-0.5ex}

Two MRH signatures, $X$ and $Y$,
are compared using an {\small $L_1$}-norm based distance measure:

\vspace{-2ex}
\begin{equation}
  d_\mathtt{raw} \left( X, Y \right)
  =
  \left\| X - Y \right\|_1
\label{eqn:raw_dist}
\end{equation}


\noindent
A decision on whether $X$ and $Y$ represent the same person (i.e.,~matched pair)
or two different persons (i.e.,~mismatched pair)
can be obtained by comparing $d_\mathtt{raw}(X,Y)$ to a threshold.
However, in order to provide further robustness to varying image conditions present in {\small $X$} and {\small $Y$},
a normalised distance can be obtained by adapting the cohort normalisation approach
originally used in speech processing~\cite{Sanderson2009ICB,Furui_PRL_1997}:

\vspace{-4ex}
\begin{equation}
  d_\mathtt{norm} (X, Y) =
  \frac
    {
    d_\mathtt{raw}(X, Y)
    }
    {
    \frac{1}{2M}
    \sum_{i=1}^{M} \left\{ d_\mathtt{raw}(\hspace{-1pt}X, C_i\hspace{-1pt}) + d_\mathtt{raw}(Y, C_i\hspace{-1pt}) \right\}
    }
\label{eqn:norm_dist}
\end{equation}

\noindent
Here, {\small $C_i$} is the $i$-th cohort face and {\small $M$} is the number of cohorts,
with the cohort faces taken from the training set.

For probes and galleries with multiple MRH signatures,
each of the probe's $K_p$ MRH signatures are individually compared to each of the gallery's $K_g$ MRH signatures,
resulting in $K_p \times K_g$ distances.
The lower the distance, the more similar two signatures are,
thus the minimum distance is taken as the final distance between a probe-gallery video pair.
The minimum distance is then compared to a threshold to obtain the final match/mismatch decision.

An appropriate threshold can be determined using a labelled set
by looking at the value which results in the minimum amount of false positives
(matching probe and gallery identities with a distance greater than the threshold)
and false negatives
(non-matching probe and gallery identities with a distance less than the threshold).
This is also referred to as minimum error rate and used in the experiments.

\section{Experiments and Discussion}
\label{sec:experiments}
\vspace{-0.5ex}

In our experiments we used the large-scale `Mobile Biometry' (MOBIO) dataset,
which has been created as part of a European project focusing
on biometric person recognition from portable devices~\cite{MOBIO2010ICPR}.
The dataset is split into three distinct sets:
one for training, one for development and one for testing.
No persons are shared across any of the three sets.

The protocol for enrolling and testing is the same
for the the development set and the test set.
There are five enrolment videos
for each user and 75 test client (positive sample) videos
for each user (15 from each session).
When producing impostor 
\linebreak
scores all the other clients are used,
for instance if in total there were 50 clients then the other
49 clients would perform an impostor attack.

For the development set, there are 20 female and 27 male users,
which results in 30,000 probe to user comparisons for females and 54,675 for males.
For the test set, there are 22 female and 39 male users,
resulting in 36,300 comparisons for females and 114,075 for males.

The MOBIO experiment protocol involves evaluating a face recognition system
on the development and test subsets for males and females.
In this paper, we present the results of the four subsets with minimum error rate (MER),
given by:

\vspace{-2ex}
\begin{equation}
  \operatorname{MER} = \min_{t} \frac{1}{2} \left( \operatorname{FAR}_t + \operatorname{FRR}_t \right)
  \label{equ:MER}
\end{equation}

\noindent
where $\operatorname{FAR}_t$ and $\operatorname{FRR}_t$
are the false acceptance rate and false rejection rate obtained at threshold~$t$.
An~equal weighting was chosen for FAR and FRR to remain application neutral.
MER is a variant of the equal error rate (EER)~\cite{Furui_PRL_1997},
but is considered to be more reliable as it does not make any assumptions about the shape of the FAR and FRR curves.

\subsection{Face Selection}
\label{subsec:face_selection}
\vspace{-0.5ex}

In the face selection approach, a subset of faces are chosen for recognition based on a particular selection metric
to characterise whether different metrics can improve recognition performance, and if so, by how much.

Our first experiment compares the recognition accuracy across the following three selection methods,
on an increasing number of faces selected:
(i)~face detection confidence,
(ii)~random selection,
and
(iii) sequential selection.
After the face are selected, the average MRH signature over all selected faces is used for recognition.
The MER results are presented in column~(a) of Fig.~\ref{fig:faceselection}.
The following three main observations can be made:

{\setlength{\leftmargini}{3ex} 
\begin{enumerate}

\item
Using multiple faces always performs better than using only one face,
but using all faces does not guarantee the best performance.
This implies that average MRH signatures are generally a good representation
of the varied samples as the performance after averaging is always better than single face.

\item
The face selection method itself affects the recognition rate drastically.
Random selection seems to provide slightly better performance
in terms of minimal error rate for recognition when compared to sequential sampling of faces.
The reason from an information point of view is that sequential faces are very likely
to have much less variation compared to faces sampled randomly throughout the video.
Face confidence, the most computationally expensive one tested,
typically gives better performance overall compared to the other two metrics.
The reason might be due to better alignment (i.e., a more frontally aligned face)
as the confidence is related to how well facial landmarks are located within the face.

\item
The optimal number of faces (the number which gives the lowest error)
varies drastically across face selection methods as well as the MOBIO subsets.
Typically, training data is used for setting parameters such as the number of faces to use,
and is assumed to have similar characteristics as test data.
We can see that even within the same dataset such as MOBIO, this assumption does not hold true.
This highlights the fragility of face selection --
the application depends on heuristic methods (i.e., number of faces or threshold of confidence)
which is very dependent on the data and method.
As such, face selection is not likely to translate well across various datasets.

\end{enumerate}

\subsection{Face Feature Clustering}
\vspace{-0.5ex}

In the cases where computation time is less of a limitation,
such as offline or batch processing,
there is potential to utilise information from all faces in a video.
We propose clustering of facial features as a pre-cursor to face recognition
to make the face matching stage computationally tractable and more memory efficient
(just storing and comparing the cluster centroids).
In the MRH framework, clustering also takes advantage of the observation made in Section~\ref{subsec:face_selection},
where higher recognition accuracy was achieved by using multiple faces rather than a single face.
This suggests average MRH signatures (such as a centroid of an MRH cluster) would provide better signatures for recognition.

The experiments for $k$-means clustering were done using single video
(where faces from just one video of a person are clustered)
and multiple videos (where faces from all videos of the same person are clustered).
The recognition results for both cases using varying $k$ are presented in column~(b) of Fig.~\ref{fig:faceselection}.
The following three main observations can be made:


\begin{enumerate}

\item
For every subset, the optimal $k$ was greater than 1.
As an example, the female development subset seems to give the best results for clustering at $k=2$.
Fig.~\ref{fig:kmeans_K002} shows a few images from a video in that subset for two clusters.
As can be observed in Fig.~\ref{fig:kmeans_K002},
clustering yielded visually discernable differences between the images.
Cluster~2 has more closely cropped faces (with borders cutting off the edges of the face),
whereas Cluster~1 shows more of the chin, hair and a bit of background.
This is reflective of face alignment errors due to inaccuracies in eye localisation
in the first stage of the video-based recognition system,
and indicates that clustering may be a good way to minimise errors introduced in earlier stages of the system.

\item
The clustering of faces from multiple videos was nearly always better
(in terms of finding the overall minimum error rate for a subset)
than clustering of faces from a single video alone.
In MOBIO, each client gallery consisted of five separate videos.
Based on the clustering results, these videos seem to be recorded in similar environments.
Due to the similarity overlap, clustering by gallery likely resulted in better performance
due to more samples in each cluster to provide more robust MRH signatures.

\item
The optimal $k$ varied depending on the MOBIO subset.
However, unlike face selection where heuristics are used to select the optimal number of faces,
there is extensive literature on clustering methods which find the `natural clusters' to fit the data~\cite{Jain10PRL}.
Thus clustering can be more robust in terms of maintaining optimal recognition accuracy
for a video recognition system across many different datasets.


\end{enumerate}

\begin{figure*}[!tb]
  \begin{minipage}{1.0\textwidth}
    \centering
    \begin{minipage}{0.01\textwidth}{\footnotesize [a]}\end{minipage}
    \begin{minipage}{0.24\textwidth}\includegraphics[width=1.15\textwidth,height=0.97\textwidth]{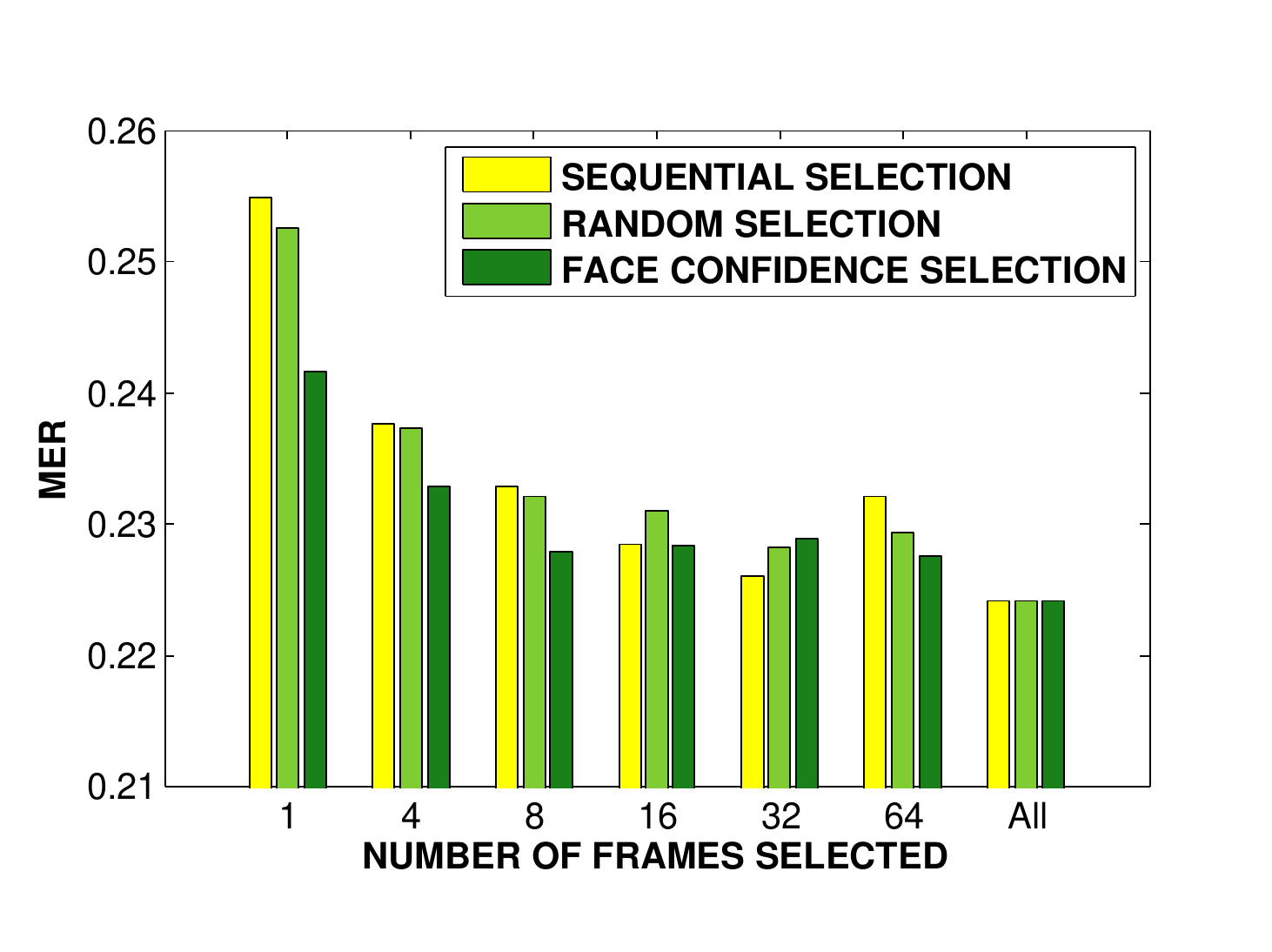}\end{minipage}
    \begin{minipage}{0.24\textwidth}\includegraphics[width=1.15\textwidth,height=0.97\textwidth]{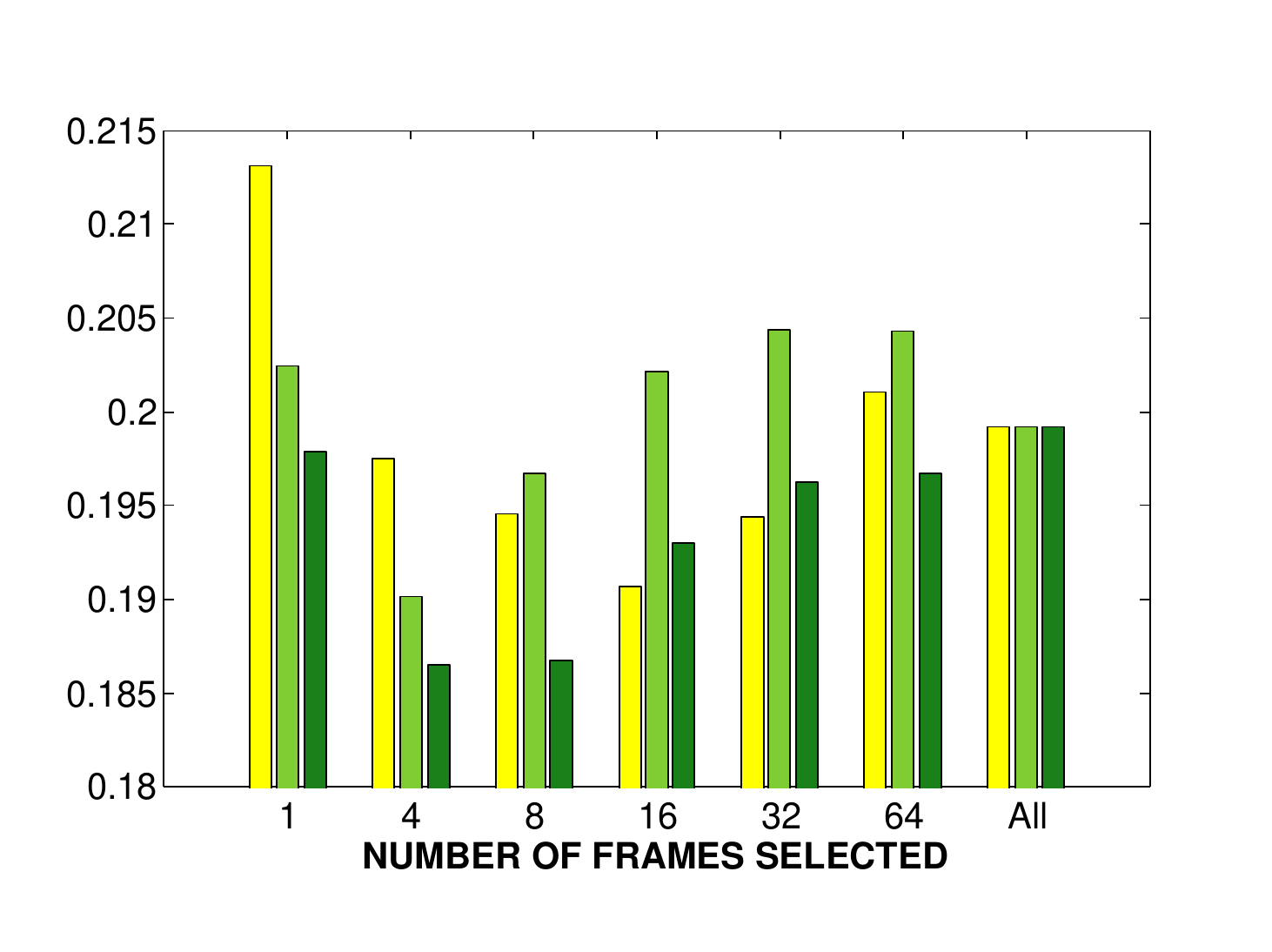}\end{minipage}
    \begin{minipage}{0.24\textwidth}\includegraphics[width=1.15\textwidth,height=0.97\textwidth]{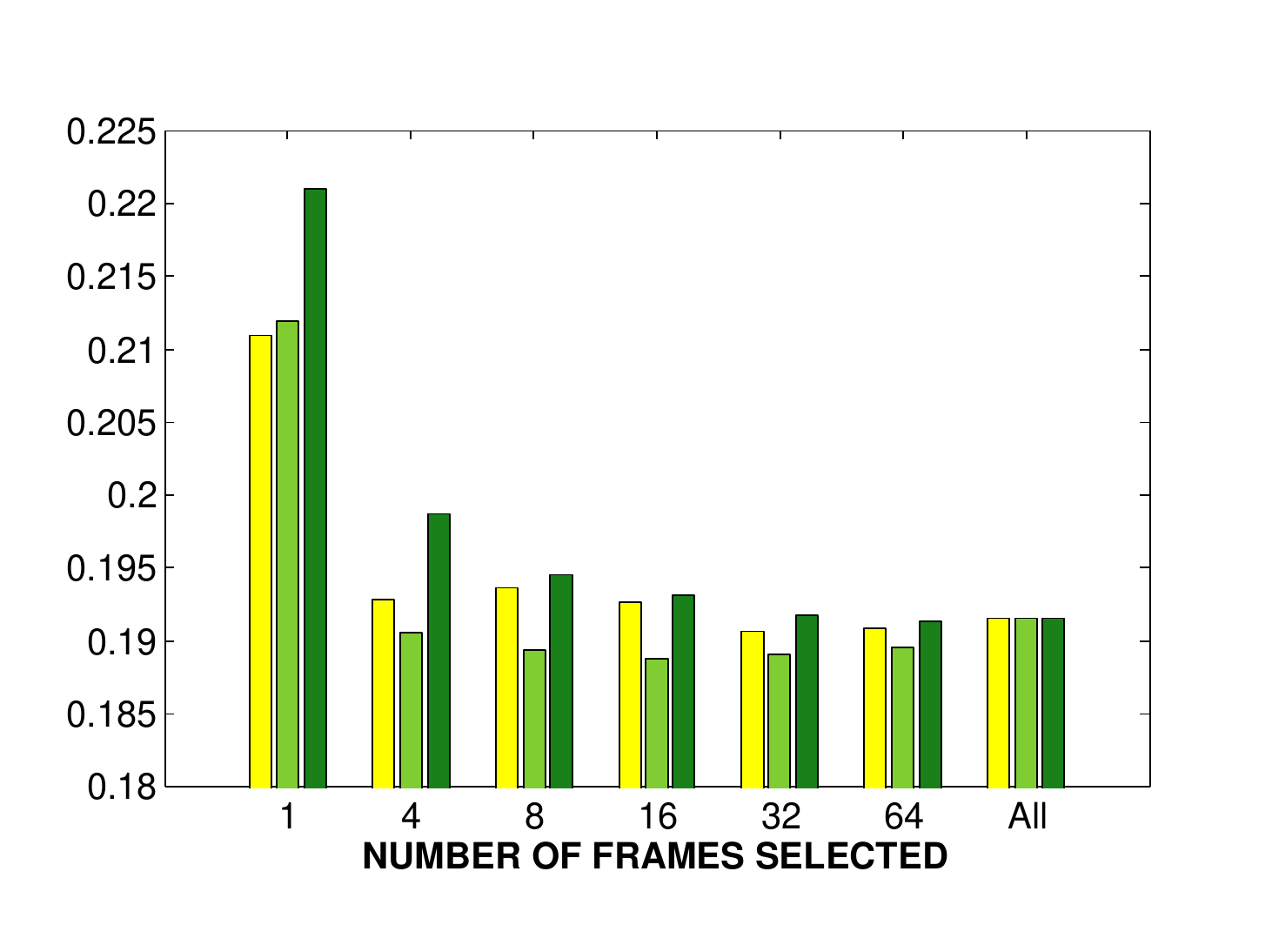}\end{minipage}
    \begin{minipage}{0.24\textwidth}\includegraphics[width=1.15\textwidth,height=0.97\textwidth]{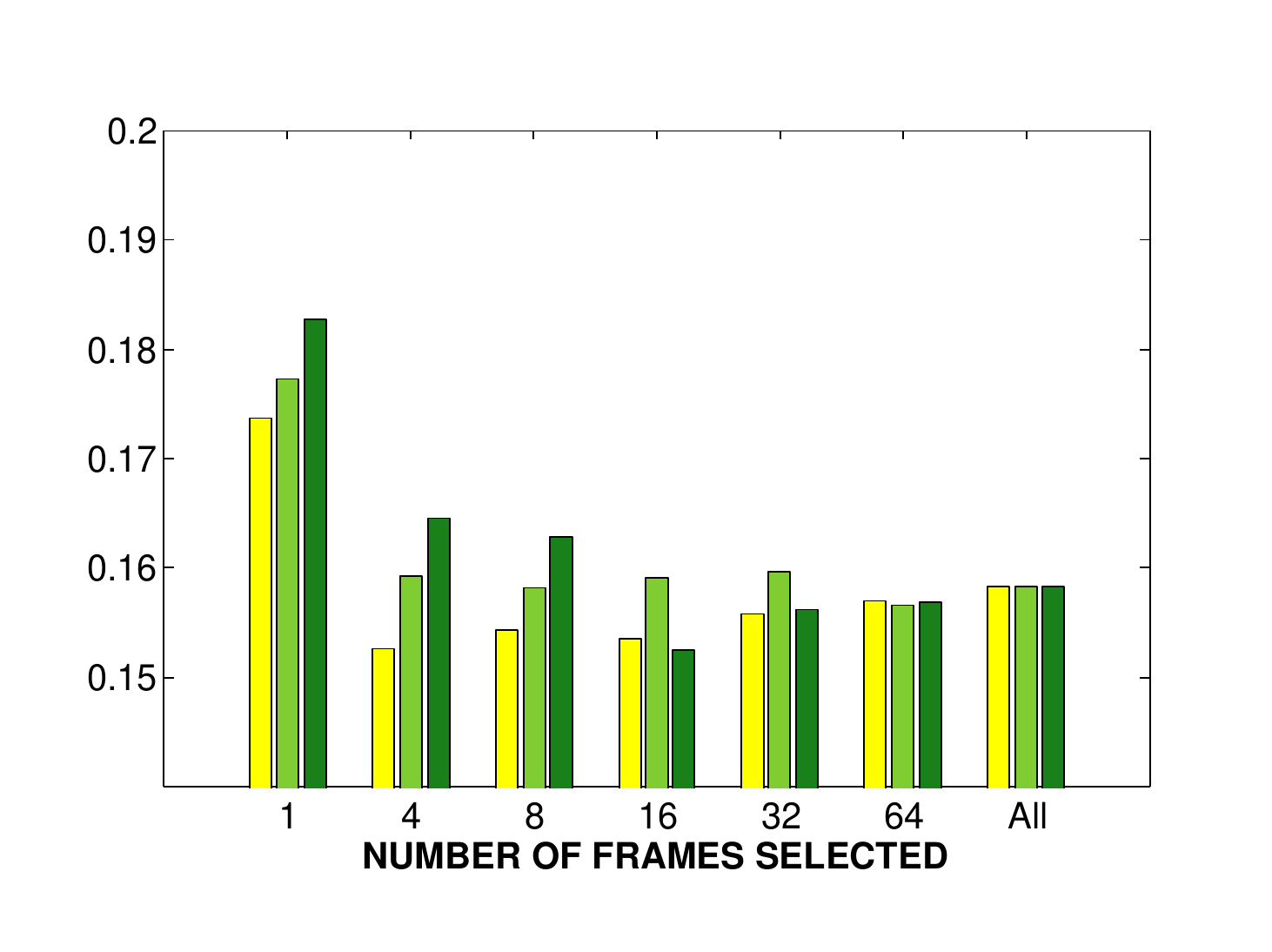}\end{minipage}
  \end{minipage}
  \begin{minipage}{1.0\textwidth}
    \centering
    \begin{minipage}{0.01\textwidth}{\footnotesize [b]}\end{minipage}
    \begin{minipage}{0.24\textwidth}\includegraphics[width=1.15\textwidth,height=0.97\textwidth]{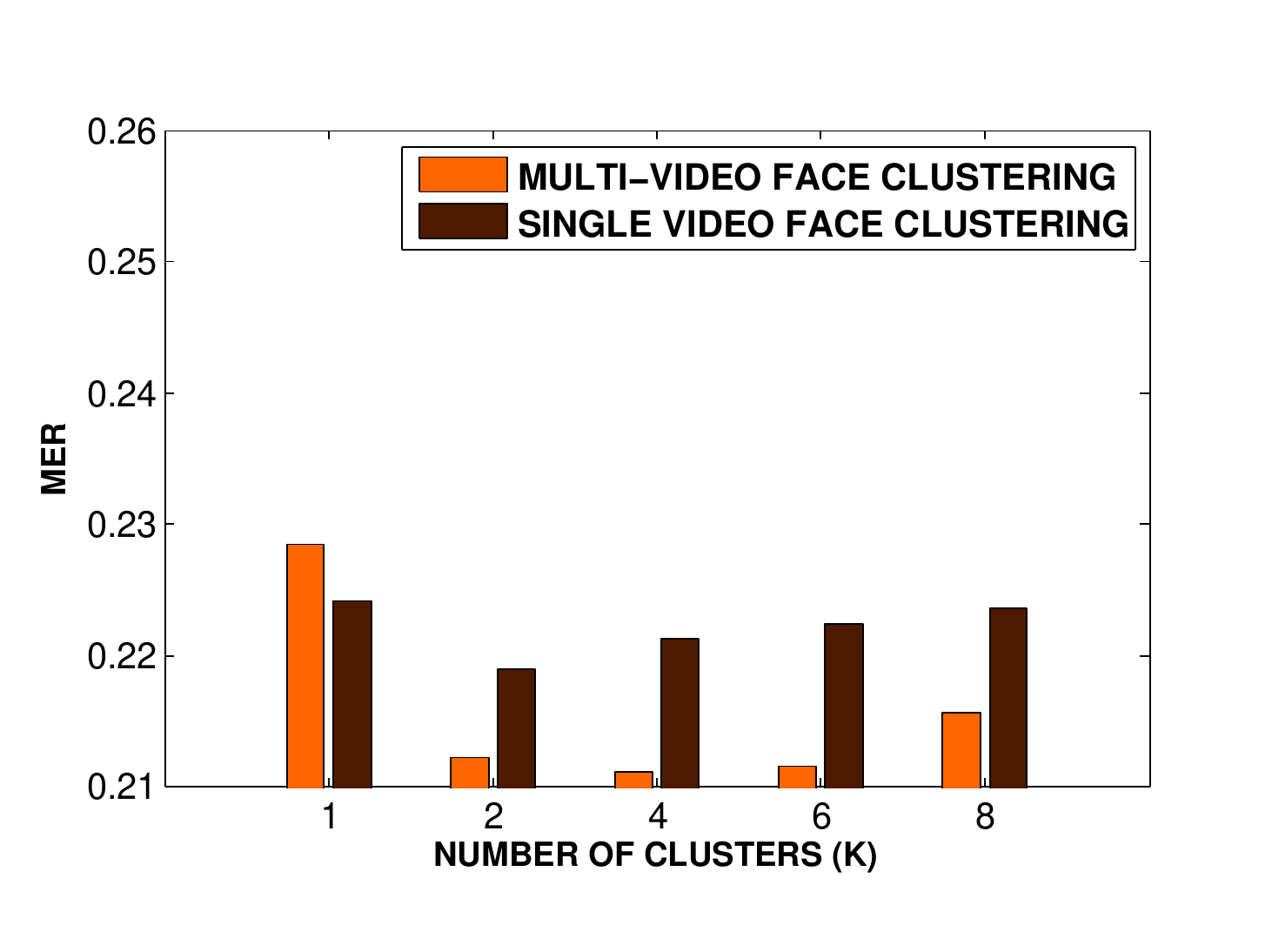}\end{minipage}
    \begin{minipage}{0.24\textwidth}\includegraphics[width=1.15\textwidth,height=0.97\textwidth]{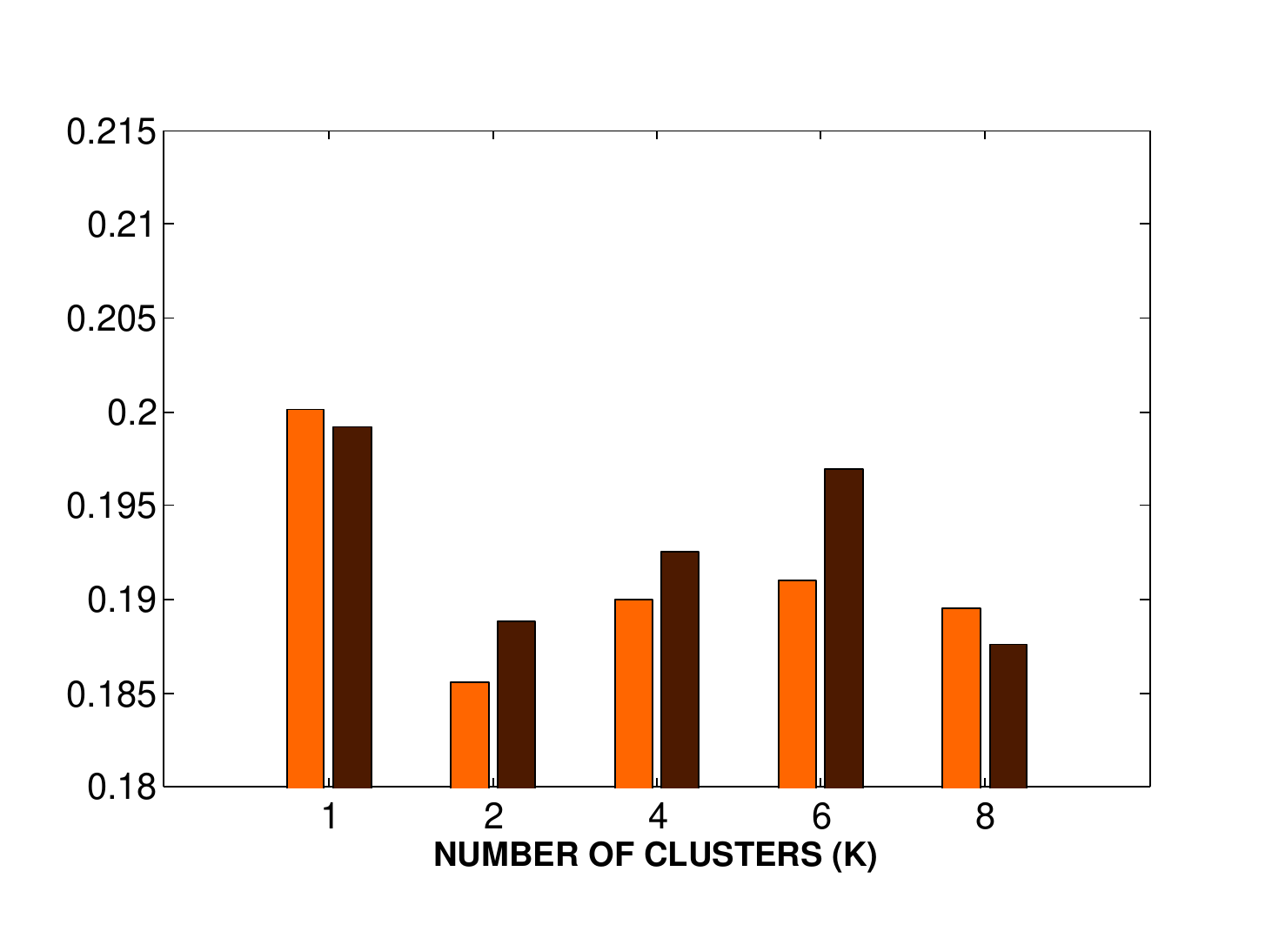}\end{minipage}
    \begin{minipage}{0.24\textwidth}\includegraphics[width=1.15\textwidth,height=0.97\textwidth]{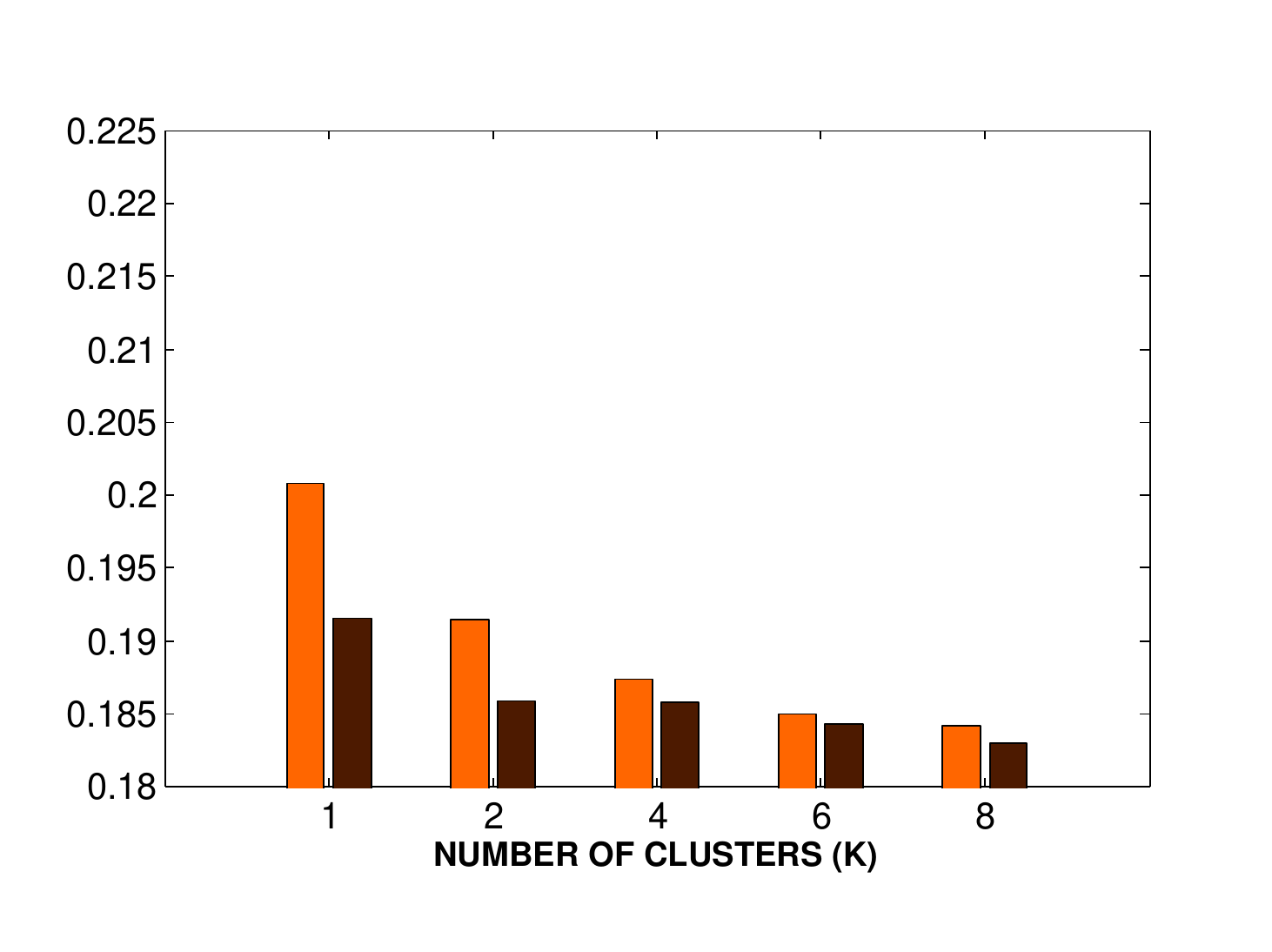}\end{minipage}
    \begin{minipage}{0.24\textwidth}\includegraphics[width=1.15\textwidth,height=0.97\textwidth]{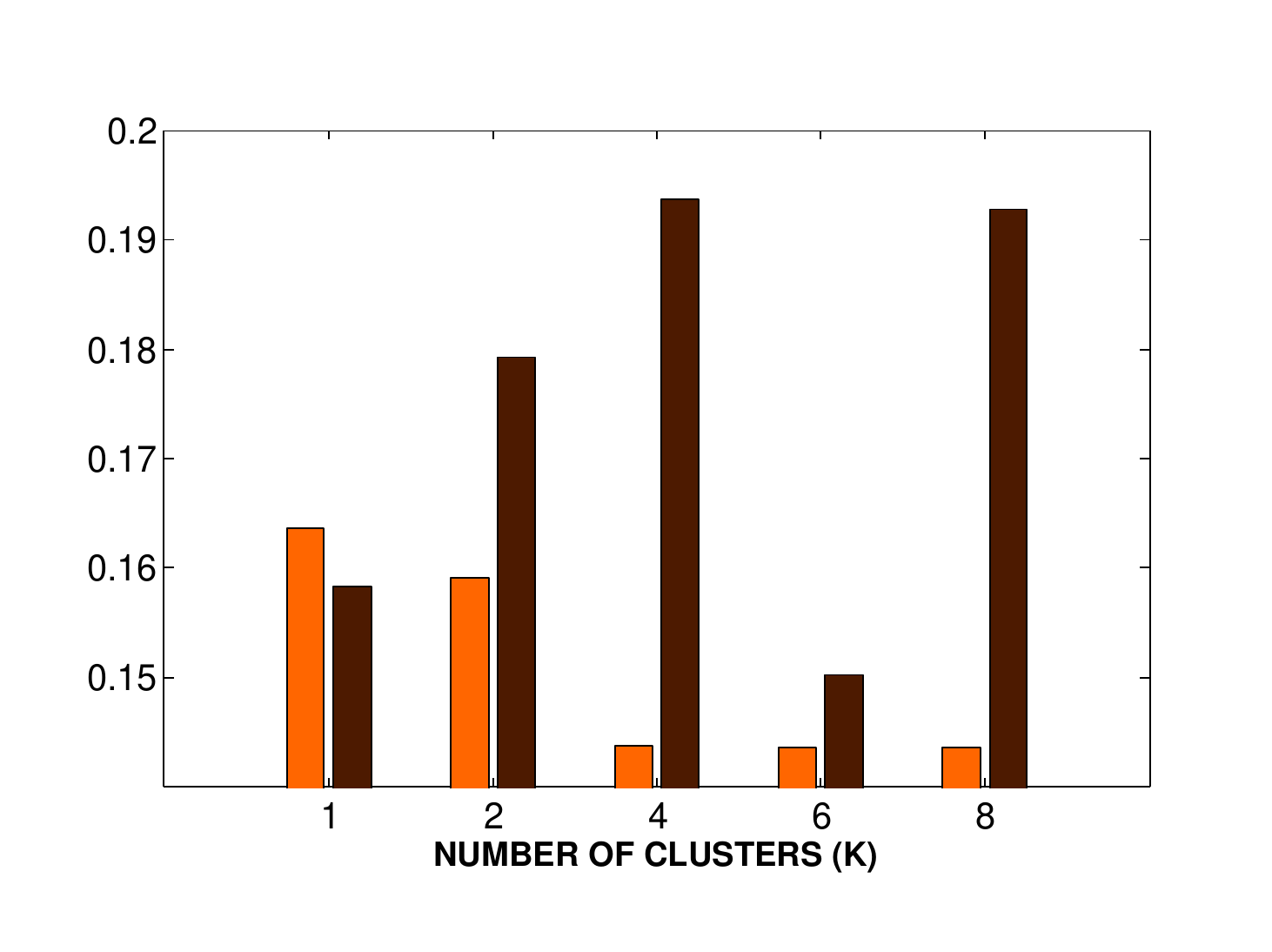}\end{minipage}
  \end{minipage}
  \begin{minipage}{1.0\textwidth}
    \centering
    \begin{minipage}{0.01\textwidth}{\footnotesize ~~~}\end{minipage}
    \begin{minipage}{0.24\textwidth}\centering\small ~~~~~~~MD\end{minipage}
    \begin{minipage}{0.24\textwidth}\centering\small ~~~~~~~FD\end{minipage}
    \begin{minipage}{0.24\textwidth}\centering\small ~~~~~~~MT\end{minipage}
    \begin{minipage}{0.24\textwidth}\centering\small ~~~~~~~FT\end{minipage}
  \end{minipage}

  \vspace{-1ex}
  \caption
    {
    \small
    MER obtained on MOBIO using:
    {\bf [a]}~three frame selection methods,
    and
    {\bf [b]}~feature clustering across single and multiple videos.
    The four subsets of MOBIO were evaluated separately:
    male development~(MD),
    female development~(FD), 
    male test~(MT), 
    female test~(FT).
    Observations for frame selection:
    (i)~multiple frames outperform one frame,
    (ii)~confidence-based selection generally gives the lowest error,
    (iii)~the number of frames for lowest error varies across subsets and selection metrics.
    Observations for feature clustering:
    (i)~the optimal $k$ is greater than 1,
    (ii)~clustering of faces from multiple videos is generally better than clustering of faces for each video independently,
    (iii)~the optimal $k$ varies depending on subset and method.
    Comparing face selection to clustering (row-by-row), 
    clustering always gives the lowest overall error.
    }
\label{fig:faceselection}
\end{figure*}

\definecolor{MyGray}{rgb}{0.9,0.9,0.9}

\begin{figure*}[!tb]
  \centering
  ~
  
  ~
  
  \colorbox{MyGray}
    {
    \begin{minipage}{0.45\textwidth}
      \centering
      \begin{minipage}{\textwidth}
        \centering
        \includegraphics[]{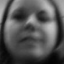}
        \includegraphics[]{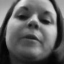}
        \includegraphics[]{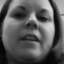}
      \end{minipage}
      \begin{minipage}{0.08\textwidth}
        \centering
        {\small cluster~1}
      \end{minipage}
    \end{minipage}
    \hfill
    }
  \colorbox{MyGray}
    {
    \begin{minipage}{0.45\textwidth}
      \centering
      \begin{minipage}{\textwidth}
        \centering
        \includegraphics[]{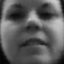}
        \includegraphics[]{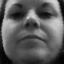}
        \includegraphics[]{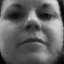}
       \end{minipage}
      \begin{minipage}{\textwidth}
        \centering
        {\small cluster~2}
      \end{minipage}
    \end{minipage}
    }
  \caption
    {
    Results from clustering of faces from a video in the female development set, using $k=2$. 
    Clustering yielded visually observable differences of two distinct face sizes resulting from inaccuracies in earlier eye localisation.
    }
  \label{fig:kmeans_K002}
\end{figure*}

\subsection{Comparing Face Selection \& Clustering}
\label{subsec:selection_clustering_compared}
\vspace{-0.5ex}

While face selection and feature clustering are not mutually exclusive components in a video-based recognition system,
both can separately contribute to reducing the computational effort to different degrees --
face selection reduces the computational requirements for the subsequent steps of feature extraction and matching (distance calculation),
while feature clustering reduces the computation requirements for the matching step only.

The computation times for each step of the face recognition process is provided in Table~\ref{tbl:compare_comp_time}.
Feature extraction is the most computationally expensive part, taking 0.390 seconds per face.
The time taken scales linearly with the number of faces selected or desired ($N$).
The calculation of the normalised distance for a pair of MRH signatures takes approximately 0.002 seconds.
However, if no clustering is performed and the distance is calculated naively on a pairwise basis per face per video,
the distance time scales to $N^2$ and can quickly exceed the computation time for feature extraction. 

For the experimental system (Table~\ref{tbl:compare_comp_time}), 
the number of faces ($N$) at which the naive matching exceeds the feature extraction time
can be found by solving $0.390*N = 0.002*N^2$,  which is $N=195$.
On the otherhand, when clustering is used, the distance scales to the number of clusters ($K$) squared, 
where $K$ will always be less than or equal to $N$.
This demonstrates how computationally inefficient it is to not use clustering or some other modeling method for reduction of signature sets.


In terms of performance for face recognition accuracy,
Table~\ref{tbl:compare_selection_cluster} shows that 
utilising information from all faces through clustering
consistently shows better accuracy than using a subset of faces.

\begin{table*}[!tb]
  \centering
  \begin{minipage}{0.45\textwidth}
    \centering
    \small
    \begin{tabular}{| r | c |}
      \hline
      {\bf Face Selection}       & {\bf Time (sec)}    \\ 
      {\bf Method}               & {\bf per 200 faces} \\ 
      \hline
      {\bf Sequential/Temporal}  & $<0.001$ \\
      {\bf Random}               & $<0.001$ \\
      {\bf Confidence}           & $16.250$ \\   
      {\bf None (all faces)}     & $<0.001$ \\
      \hline
    \end{tabular}
  \end{minipage}
  \begin{minipage}{0.45\textwidth}
    \centering
    \small
    \begin{tabular}{| c | c |}
      \hline
      {\bf Number of}    & {\bf Time (sec)}    \\ 
      {\bf Clusters (K)} & {\bf per 200 faces} \\ 
      \hline
      {\bf 1} & $0.003$ \\
      {\bf 2} & $0.027$ \\
      {\bf 4} & $0.037$ \\   
      {\bf 6} & $0.060$ \\
      {\bf 8} & $0.125$ \\
      \hline
    \end{tabular}
  \end{minipage}
  \caption
    {
    Time taken on an Intel Xeon CPU @ 3.0 GHz system, using Linux 2.6.24 and GCC 4.2.4.
    Face selection time scales linearly to the number of original input faces.
    For the $N$ selected faces, 
    clustering time scales linearly to $NK$ where $K$ is number of clusters
    (maximum number of iterations is limited to 20).
    }
  \label{tbl:compare_comp_time}
\end{table*}

\begin{table*}[!tb]
  \centering
  \small
  ~
  
  ~
  
  \begin{tabular}{| r | c | c | c || c | c | c |}
    \cline{2-7}
    \multicolumn{1}{l|}{}      & \multicolumn{3}{c||}{\bf ~Best Face Selection~}&  \multicolumn{3}{c|}{\bf ~Best Feature Clustering~~} \\ \cline{2-7}
    \multicolumn{1}{l|}{}      & {\bf ~Method~} & {\bf ~Faces~} & {\bf ~MER~}   &  {\bf ~Method~}  & {\it\bf ~k~} & {\bf ~MER~}       \\ \hline
	  {\bf ~Male dev subset~}    & any            &  all          & $22.41$       &  multiple        & $4$          & $21.12$           \\ 
	  {\bf ~Female dev subset~}  & face conf.     &  $4$          & $18.65$       &  multiple        & $2$          & $18.56$           \\ \cline{1-7}
	  {\bf ~Male test subset~}   & random         &  $16$         & $18.88$       &  single          & $8$          & $18.30$           \\ 
	  {\bf ~Female test subset~} & face conf.     &  $16$         & $15.25$       &  multiple        & $4$          & $14.36$           \\ \hline
  \end{tabular}
  \vspace{1ex}
  \caption
    {
    The lowest Minimum Error Rate (MER \%) for face selection and feature clustering on MOBIO.
    Under face selection methods,
    `any' refers to any of the three methods tested,
    `face conf.' refers to face detection confidence.
    Under feature clustering methods,
    `single' refers to clustering faces within each video,
    `multiple' refers to clustering of faces from multiple videos within a person's gallery.
    The lowest MER for each MOBIO subset is always obtained through clustering.
    }
  \label{tbl:compare_selection_cluster}
\end{table*}

As separately observed both in the face selection and feature clustering experiments,
the optimal
\linebreak
value (i.e., number of faces or clusters) varied depending on the dataset and method.
It was noted that for face selection, since the thresholds are chosen heuristically,
this approach is particularly fragile to variations between datasets,
which would lead to suboptimal performance.
For feature clustering, variations in the optimal $k$ are less of an issue
as there is extensive work in finding the `natural clusters' to fit the data~\cite{Jain10PRL}.
Thus clustering is a more robust and reliable means of consistently boosting face recognition accuracy.

The above trade-offs between computation and accuracy are interesting to characterise
as they aid in determining which approach is most suitable for particular applications.
For example, face selection is more suitable for systems which have real-time requirements
(such as live video monitoring) or limited computation restrictions (such as mobile phones).
In contrast, feature clustering is more suitable for batch or offline processing
such as forensic applications and pre-processed galleries for watchlists.

\section{Conclusions and Future Work}
\label{sec:conclusion}

In this paper, we examined two approaches of improving the performance of a video-based face recognition system
--- face selection and face feature clustering.
Three methods of face selection were investigated:
face detection confidence, random selection and sequential selection.

In comparing the three selection methods, it was found that:
{\bf (i)}~using multiple faces is always better than using a single face alone,
{\bf (ii)}~the face detection confidence metric typically provides better results when using a subset of faces,
and
{\bf (iii)}~the optimal number of faces to use varies drastically across selection methods and datasets (subsets of MOBIO).

For feature clustering, we used a $k$-means approach and found that the optimal $k$ varied across datasets,
and that more faces provided better cluster representations for recognition
(i.e., clustering faces from multiple videos together is better than clustering from a single video).

When compared to face selection, the lowest error rates were always obtained through clustering
at the expense of higher computational effort.
With face selection, the computation of a selection metric is typically low.
However, its major drawback is that the selection of the number of faces is done in a heuristic manner,
and as such highly dependent on both dataset and face selection metric.
The parameters of face selection are not likely to translate well across datasets, thus potentially giving sub-optimal results.

With face feature clustering, the optimal number of clusters may vary across datasets,
however there are many principled methods of adaptive clustering to find the optimal number of clusters~\cite{Jain10PRL}.
As such, the clustering approach is more robust and transferable across datasets.
However, its main drawback is the computational effort required
for face detection in all frames and the subsequent feature extraction.

Based on the above trade-offs, our experiments suggest that designers of video-based recognition systems
should use facial feature clustering if they are able to process videos in a batch fashion (offline),
as clustering can robustly maximise recognition accuracy.
This would also be applicable to galleries of online systems as the galleries are typically processed in batch.
In contrast, if the application has real-time requirements such as live video monitoring in surveillance,
the selection of faces using a good face confidence metric may make the most sense.

Though we investigated face selection and face feature clustering individually,
it is still worth exploring the combination of these two techniques, such as clustering on selected faces. 
In addition, it would also be worthwhile to further examine face feature clustering on other face recognition techniques.

\balance

\section{Acknowledgements}

NICTA is funded by the Australian Government's {\it Department of Broadband, Communications and Digital Economy}
as well as the Australian Research Council through {\it Backing Australia's Ability}
and the {\it ICT Research Centre of Excellence} programs.
NICTA's Queensland Laboratory is in part funded by the Queensland State Government.

This work is financially supported by the Australian Government
through the National Security Science and Technology Branch within
the Department of the Prime Minister and Cabinet.  This support 
does not represent and endorsement of the contents or conclusions
of the project.

\small
\bibliographystyle{IEEEtran}
\bibliography{video_recognition}

\begin{thebibliography}{10}
\providecommand{\url}[1]{#1}
\csname url@samestyle\endcsname
\providecommand{\newblock}{\relax}
\providecommand{\bibinfo}[2]{#2}
\providecommand{\BIBentrySTDinterwordspacing}{\spaceskip=0pt\relax}
\providecommand{\BIBentryALTinterwordstretchfactor}{4}
\providecommand{\BIBentryALTinterwordspacing}{\spaceskip=\fontdimen2\font plus
\BIBentryALTinterwordstretchfactor\fontdimen3\font minus
  \fontdimen4\font\relax}
\providecommand{\BIBforeignlanguage}[2]{{%
\expandafter\ifx\csname l@#1\endcsname\relax
\typeout{** WARNING: IEEEtran.bst: No hyphenation pattern has been}%
\typeout{** loaded for the language `#1'. Using the pattern for}%
\typeout{** the default language instead.}%
\else
\language=\csname l@#1\endcsname
\fi
#2}}
\providecommand{\BIBdecl}{\relax}
\BIBdecl

\bibitem{Gary-ICCV-2006}
G.~B. Huang, V.~Jain, and E.~Learned-Miller, ``Unsupervised joint alignment of
  complex images,'' in \emph{International Conference on Computer Vision
  (ICCV)}, 2007.

\bibitem{Sanderson2009ICB}
C.~Sanderson and B.~C. Lovell, ``Multi-region probabilistic histograms for
  robust and scalable identity inference,'' in \emph{International Conference
  on Biometrics, Lecture Notes in Computer Science (LNCS)}, vol. 5558, 2009,
  pp. 199--208.

\bibitem{Matta2009JVLC}
F.~Matta and J.-L. Dugelay, ``Person recognition using facial video
  information: A state of the art,'' \emph{Journal of Visual Languages and
  Computing}, vol.~20, pp. 180--187, 2009.

\bibitem{Wang2009WASET}
H.~Wang, Y.~Wang, and Y.~Cao, ``Video-based face recognition: A survey,''
  \emph{World Academy of Science, Engineering and Technology}, vol.~60, pp.
  293--302, 2009.

\bibitem{Zhao03ACMCS}
W.~Zhao, R.~Chellappa, P.~J. Phillips, and A.~Rosenfeld, ``Face recognition: A
  literature survey,'' \emph{ACM Comput. Surv.}, vol.~35, no.~4, pp. 399--458,
  2003.

\bibitem{Shan10SCI}
C.~Shan, ``Face recognition and retrieval in video,'' \emph{Studies in
  Computational Intelligence}, vol. 287, pp. 235--260, 2010.

\bibitem{MOBIO2010ICPR}
S.~Marcel, P.~M. C.~McCool, T.~Ahonen, and J.~Cernocky, ``{M}obile {B}iometry
  (\mbox{MOBIO}) {F}ace and {S}peaker {V}erification {E}valuation,'' Martigny,
  Switzerland, {IDIAP} {R}esearch {R}eport {RR}-09-2010, 2010.

\bibitem{Gorodnichy02FG}
D.~Gorodnichy, ``On importance of nose for face tracking,'' in \emph{In
  proceedings of IEEE International Conference on Automatic Face and Gesture
  Recognition (FG)}, 2002, pp. 181--186.

\bibitem{Villegas08CVPR}
M.~Villegas and R.~Paredes, ``Simultaneous learning of a discriminative
  projection and prototypes for nearest-neighbor classification,'' in
  \emph{Proc. IEEE Conf. Computer Vision and Pattern Recognition (CVPR)}, 2008.

\bibitem{Berrani05AVSS}
S.~Berrani and C.~Garcia, ``Enhancing face recognition from video sequences
  using robust statistics,'' in \emph{In proceedings of IEEE International
  Conference on Advanced Video and Signal-Based Surveillance (AVSS)}, 2005, pp.
  324--329.

\bibitem{Lee03CVPR}
K.~Lee, J.~Ho, M.~Yang, and D.~Kriegman, ``Video-based face recognition using
  probabilistic appearance manifolds,'' in \emph{In proceedings of IEEE
  Conference on Computer Vision and Pattern Recognition (CVPR)}, 2003, pp.
  313--320.

\bibitem{Arandjelovi05CVPR}
O.~Arandjelovic, G.~Shakhnarovich, G.~Fisher, J.~Cipolla, and R.~Zisserman,
  ``Face recognition with image sets using manifold density divergence,'' in
  \emph{In proceedings of IEEE Conference on Computer Vision and Pattern
  Recognition (CVPR)}, vol.~1, 2005, pp. 581--588.

\bibitem{Arandjelovi06ICPR}
O.~Arandjelovic and R.~Cipolla, ``Face set classification using maximally
  probable mutual modes,'' in \emph{In proceedings International Conference on
  Pattern Recognition (ICPR)}, vol.~1, 2006, pp. 511--514.

\bibitem{Viola2001IJCV}
P.~A. Viola and M.~J. Jones, ``Robust real-time face detection,''
  \emph{International Journal of Computer Vision}, vol.~57, no.~2, pp.
  137--154, 2004.

\bibitem{Nowak_ECCV_2006}
E.~Nowak, F.~Jurie, and B.~Triggs, ``Sampling strategies for bag-of-features
  image classification,'' in \emph{European Conf. Computer Vision (ECCV), Part
  IV, Lecture Notes in Computer Science (LNCS)}, vol. 3954, 2006, pp. 490--503.

\bibitem{Gonzales_2007}
R.~Gonzales and R.~Woods, \emph{Digital Image Processing}, 3rd~ed.\hskip 1em
  plus 0.5em minus 0.4em\relax Prentice Hall, 2007.

\bibitem{Jain10PRL}
A.~K. Jain, ``Data clustering: 50 years beyond {K}-means,'' \emph{Pattern
  Recognition Letters}, vol.~31, no.~8, pp. 651--666, June 2010.

\bibitem{Mackay2003}
D.~MacKay, \emph{Information Theory, Inference, and Learning Algorithms}.\hskip
  1em plus 0.5em minus 0.4em\relax Cambridge University Press, 2003.

\bibitem{Furui_PRL_1997}
S.~Furui, ``Recent advances in speaker recognition,'' \emph{Pattern Recognition
  Letters}, vol.~18, no.~9, pp. 859--872, 1997.

\end{thebibliography}

\end{document}